\documentclass[lettersize,journal]{IEEEtran}
\usepackage{amsmath,amsfonts}
\usepackage{algorithmic}
\usepackage{algorithm}
\usepackage{array}
\usepackage[caption=false,font=normalsize,labelfont=sf,textfont=sf]{subfig}
\usepackage{textcomp}
\usepackage{stfloats}
\usepackage{url}
\usepackage{verbatim}
\usepackage{graphicx}
\usepackage{cite}

\usepackage{threeparttable}

\usepackage{xcolor}

\usepackage{multirow}

\usepackage[numbers,sort&compress]{natbib}

\usepackage[figuresleft]{rotating}

\begin{document}

\title{Milestones in Autonomous Driving and Intelligent Vehicles: Survey of Surveys}

\author{Long Chen,~\IEEEmembership{Senior Member,~IEEE,} Yuchen Li, Chao Huang, Bai Li, Yang Xing, 

Daxin Tian,~\IEEEmembership{Senior Member,~IEEE,} Li Li,~\IEEEmembership{Fellow,~IEEE,} Zhongxu Hu, Xiaoxiang Na, Zixuan Li,

Siyu Teng, Chen Lv,~\IEEEmembership{Senior Member,~IEEE,} Jinjun Wang, Dongpu Cao,~\IEEEmembership{Senior Member,~IEEE,} 

Nanning Zheng,~\IEEEmembership{Fellow,~IEEE} and Fei-Yue Wang,~\IEEEmembership{Fellow,~IEEE} 
\thanks{
Manuscript received Sep 29, 2022; accepted Nov 01, 2022. Date of publication *** **, 20xx; date of current version *** **, 20xx. (Corresponding author: Fei-Yue Wang.)

This work is supported by the National Natural Science Foundation of China (62006256) and Key Research and Development Program of Guangzhou (202007050002  202007050004).

Long Chen and Fei-Yue Wang are with the State Key Laboratory of Management and Control for Complex Systems, Institute of Automation, Chinese Academy of Sciences, Beijing, 100190, China, and Long Chen is also with Waytous Ltd. (e-mail: long.chen@ia.ac.cn; feiyue.wang@ia.ac.cn). 

Yuchen Li and Siyu Teng are with BNU-HKBU United International College, Zhuhai, 519087, China and Hong Kong Baptist University, Kowloon, Hong Kong, 999077, China (e-mail: liyuchen2016@hotmail.com; siyuteng@ieee.org).

Chao Huang is with the Department of Industrial and System Engineering, Hong Kong Polytechnical University (e-mail: hchao.huang@polyu.edu.hk).

Bai Li is with the College of Mechanical and Vehicle Engineering, Hunan University (e-mail: libai@zju.edu.cn). 

Yang Xing is with the School of Aerospace, Transport, and Manufacturing, Cranfield
University (e-mail: yang.x@cranfield.ac.uk). 

Daxin Tian is with the School of Transportation Science and Engineering, Beihang University (e-mail: dtian@buaa.edu.cn). 

Li Li is with the Department of Automation, Tsinghua University (e-mail: li-li@tsinghua.edu.cn). 

Zhongxu Hu and Chen Lv are with the Department of Mechanical and Aerospace Engineering of Nanyang Technological University (e-mail: Huzhongxu.hu@ntu.edu.sg; lyuchen@ntu.edu.sg). 

Xiaoxiang Na is with the Centre for Sustainable Road Freight, University of Cambridge (e-mail: xnhn2@cam.ac.uk).

Zixuan Li is with Waytous Ltd. (e-mail: lizixuan981258655@gmail.com).

Dongpu Cao is with the School of Mechanical Engineering, Tsinghua University (e-mail: dp\_cao2016@163.com).

Jinjun Wang and Nanning Zheng are with the College of Artificial Intelligence, Xi'an Jiaotong University (e-mail: jinjun@mail.xjtu.edu.cn; nnzheng@mail.xjtu.edu.cn).
}
}

\markboth{Journal of \LaTeX\ Class Files,~Vol.~14, No.~8, August~2021}%
{Shell \MakeLowercase{\textit{et al.}}: A Sample Article Using IEEEtran.cls for IEEE Journals}


\maketitle

\begin{abstract}

Interest in autonomous driving (AD) and intelligent vehicles (IVs) is growing at a rapid pace due to the convenience, safety, and economic benefits. Although a number of surveys have reviewed research achievements in this field, they are still limited in specific tasks, lack of systematic summary and research directions in the future. Here we propose a Survey of Surveys (SoS) for total technologies of AD and IVs that reviews the history, summarizes the milestones, and provides the perspectives, ethics, and future research directions. To our knowledge, this article is the first SoS with milestones in AD and IVs, which constitutes our complete research work together with two other technical surveys. We anticipate that this article will bring novel and diverse insights to researchers and abecedarians, and serve as a bridge between past and future.


\end{abstract}

\section{Introduction}

\IEEEPARstart{A}{utonomous} driving (AD) and intelligent vehicles (IVs) have recently attracted significant attention from academia as well as industry because of a range of potential benefits. Surveys on AD and IVs occupy an essential position in gathering research achievements, generalizing entire technology development, and forecasting future trends. However, a large majority of surveys only focus on the specific task, and they may have a negative impact on conducting research for abecedarians. The purpose of our work is to systematically summarize the development of AD and propose future research directions from an overall perspective. This paper is the Part 1 of the “Milestones in Autonomous Driving and Intelligent Vehicles”, a survey of surveys (SoS), and the Part 2 and 3 will be published soon. This paper collects milestones on surveys of AD and IVs and introduces research perspectives, ethics, and future directions. In other two papers, we review crucial technologies in AD including perception, planning, control, etc. We expect that our work can be considered as a bridge between past and future.

\subsection{History of Autonomous Driving \& Intelligent Vehicles}
The first automated, radio-operated vehicle was successfully tested in the USA on $5$th August $1921$. In $1953$, Radio Corporation of America (RCA) Laboratories successfully developed a miniature vehicle that was navigated and controlled by wires. The IVs or called remotely piloted vehicles were limited by technological developments and could only achieve a single unstable function.

The development of AD witnessed a breakthrough in $1980$s thanks to the development of computer technology. The US Defence Advanced Research Projects Agency (DARPA) established the Autonomous Land Vehicle (ALV) program in 1983, involving Carnegie Mellon University (CMU), Stanford University, and other academic institutions to realize AD which is the first time to integrate LiDAR, computer vision, and automated control methods. In 1989, CMU pioneered the use of neural networks to guide the control of IVs, and this development laid a foundation for intelligent control techniques.

At the beginning of the $21$ century, several competitions worldwide promoted the research on AD. Starting in $2004$, DARPA held three competitions to evaluate the capabilities of IVs in harsh and complex environments. Stanford University won the first prize in the competition in $2005$, and their vehicle was equipped with a camera, a LiDAR, a radar, a Global Positioning System (GPS), and an Intel CPU. The first Chinese ``Intelligent Vehicles Future Challenge Program" was held in $2009$, which attracted seven groups to participate, including Hunan University, Beijing Institute of Technology, Shanghai Jiaotong University, Xi'an Jiaotong University, Tsinghua University, National University of Defense Technology and University of Parma.

In $2010$s, owing to the development of neural networks as well as the computing platform, IVs have gradually moved from private roads to urban roads. VIsLab implemented cross-border transport of IVs from Parma to Shanghai. In $2016$, Drive.ai was permitted to test IVs in California. The nuTonomy in Singapore ran a number of autonomous taxis in the same year.

With regard to AD levels, the Society of Automotive Engineers (SAE) has divided AD into $6$ levels from L$0$ to L$5$. By $2030$, $82$ million IVs with L4/L5 will run in the US, Europe, and China. Although AD technology has got impressive development, the issues still exist. In addition, it is still legally in question whether an IV could undertake the responsibility when it involves a traffic accident.


\subsection{Paper Structure}
We divide the article into five sections, including introduction, overall, datasets, perspectives \& future, and conclusion. The introduction section contains a brief introduction of history of AD and our contributions. In the overall section, we category the collected survey papers and analyse the statistic results. We also summarize the dataset information on the AD in the datasets section. In the perspective \& future section, we provide research perspectives, ethics and future directions on AD.


\begin{table}[h!]
\centering
\caption{Distributions of Reviewed Surveys}
\begin{tabular}{llc}
\hline
\multicolumn{1}{c}{Article Category}        & \multicolumn{1}{c}{Concrete Theme}                     & \multicolumn{1}{c}{Number} \\ \hline
Overall                    & Overall                         & 13     \\
Perception & Localization              & 17     \\
Perception & Static Object Detection & 10     \\ 
Perception & Dynamic Object Detection & 27 \\
Perception & Scene Understanding & 3 \\
Perception & Tracking & 2 \\
Perception & Prediction & 2 \\
Planning  & Planning                  & 6      \\
Planning  & Decision-making \& End-to-End    & 2      \\ 
Control                    & Control                         & 7      \\ 
System     & System \& Platform       & 4      \\
System     & Hardware                  & 3      \\
System     & Software                  & 1      \\ 
Communication               & Communication                         & 15     \\ 
Testing    & Simulation                & 2      \\
Testing    & Interpretability                & 1      \\ 
Interaction                 & Human-Machine Interface                        & 1  \\
Scenes &Special Scenes & 6
\\ \hline
\end{tabular}
\label{table:overall_number}
\end{table}



\subsection{Contributions}


In this paper, we collect $122$ survey articles, analyse datasets, and provide research difficulties, directions for future research, and ethics in AD. The most important thing is that the research of AD and IVs has entered a bottleneck period. We wish this article could bring novel and diverse insights for researchers to make breakthroughs.

We summarize three contributions of this article: 

\quad 1. We introduce an SoS on AD and IVs. In this article, we collect the milestone surveys and category them into several sub-sections.

\quad 2. We enumerate the characteristics of AD datasets and summarize the current research perspectives, ethics, and future directions on AD.

\quad 3. We conduct a systematic study that attempts to be a bridge between past and future on AD and IVs, and this SoS is the Part 1 of our whole research. 

\section{Overall}

\begin{figure*}
\centering  
\includegraphics[width=18cm]{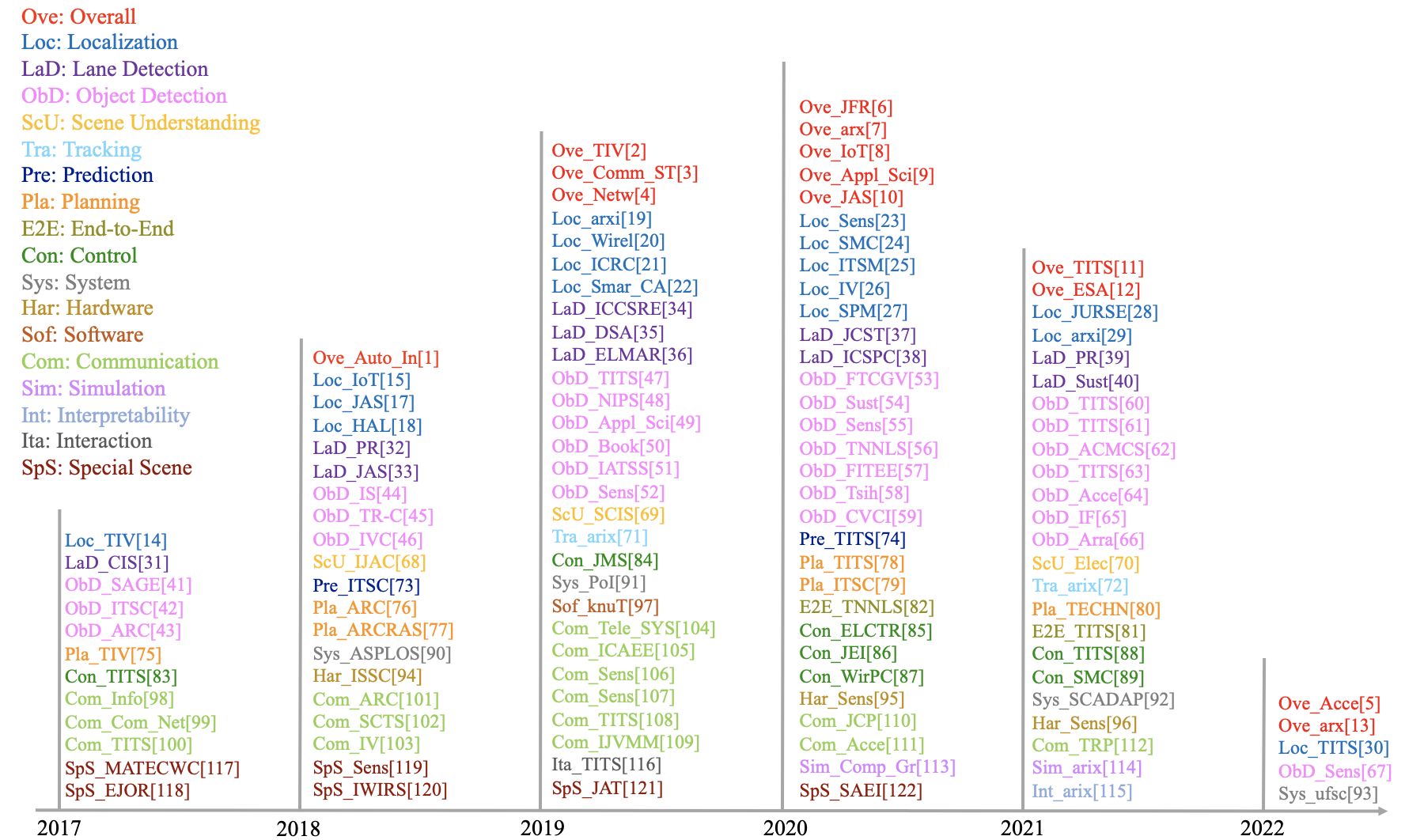}

\caption{We provide all the collected papers on the time axis with abbreviations, consisting of the categories, published journals and the serial number.}
\label{fig:whole_surveys}
\end{figure*}

We select $122$ survey articles in our paper and the Table \ref{table:overall_number} shows the categories and the corresponding numbers of papers. All the surveys are categorized into several sub-sections, including the overall \cite{survey_1, survey_2, survey_3, survey_4, survey_5, survey_6, survey_7, survey_8, survey_9, survey_10, survey_11, survey_12, survey_13}, localization \cite{survey_loca_1, survey_loca_2, survey_loca_3, survey_loca_4, survey_loca_5, survey_loca_6, survey_loca_7, survey_loca_8, survey_loca_9, survey_loca_10, survey_loca_11, survey_loca_12, survey_loca_13, survey_loca_14, survey_loca_15, survey_loca_16, survey_loca_17}, static object detection \cite{survey_static_1, survey_static_2, survey_static_3, survey_static_4, survey_static_5, survey_static_6, survey_static_7, survey_static_8, survey_static_9, survey_static_10}, dynamic object detection \cite{survey_dynamic_1, survey_dynamic_2, survey_dynamic_3, survey_dynamic_4, survey_dynamic_5, survey_dynamic_6, survey_dynamic_7, survey_dynamic_8, survey_dynamic_9, survey_dynamic_10, survey_dynamic_11, survey_dynamic_12, survey_dynamic_13, survey_dynamic_14, survey_dynamic_15, survey_dynamic_16, survey_dynamic_17, survey_dynamic_18, survey_dynamic_19, survey_dynamic_20, survey_dynamic_21, survey_dynamic_22, survey_dynamic_23, survey_dynamic_24, survey_dynamic_25, survey_dynamic_26, survey_dynamic_28}, scene understanding \cite{survey_scene_1, survey_scene_2, survey_scene_3}, tracking \cite{survey_tracking_1, survey_tracking_2}, prediction \cite{survey_prediction_1, survey_prediction_2}, planning \cite{survey_planning_1, survey_planning_2, survey_planning_3, survey_planning_4, survey_planning_5, survey_planning_6}, E2E \cite{survey_e2e_1, survey_e2e_2}, control \cite{survey_control_1, survey_control_2, survey_control_3, survey_control_4, survey_control_5, survey_control_6, survey_control_7}, system \cite{survey_system_1, survey_system_2, survey_system_4, survey_system_5}, hardware \cite{survey_hardware_1, survey_hardware_2, survey_hardware_3}, software \cite{survey_software_1},  communication \cite{survey_communication_1, survey_communication_2, survey_communication_3, survey_communication_4, survey_communication_5, survey_communication_6, survey_communication_7, survey_communication_8, survey_communication_9, survey_communication_10, survey_communication_11, survey_communication_12, survey_communication_13, survey_communication_14, survey_communication_15}, simulation \cite{survey_simulation_1, survey_simulation_2}, interpretability \cite{survey_explainable}, Human-Machine Interface (HMI) \cite{survey_human_machine_1}, and special scenes \cite{survey_special_1, survey_special_2, survey_special_3, survey_special_4, survey_special_5, survey_special_6}. Table \ref{tabel:overall_key_surveys} presents a few highly cited surveys of each sub-section. We provide the title of these articles with the categories, the number of citations, the publication year and a number of special keywords which assist researchers to find the target paper quickly. We plot the whole collected articles on a timeline as Fig. \ref{fig:whole_surveys}. Readers can clearly identify the research area and the published journal of each article according to the abbreviations, and locate the article title and other information by serial numbers. For example, ``Ove${\_}$TIV[2]" in this figure represents the article can be found at the reference list with index 2, and it belongs to the ``Overall" category and published in IEEE Transaction on Intelligent Vehicle (TIV).

\begin{sidewaystable*}[]
\centering
\caption{The crucial surveys and relative in formations of each sub-task on autonomous driving}
\begin{tabular}{p{13.5cm}cccc} \hline
\multicolumn{1}{c}{Article Name}                                                                                            & Category          & Cite & Year & Characteristics \\ \hline
A Survey of Deep Learning Techniques for Autonomous Driving \cite{survey_6}                                                                 & Overall          & \textbf{509}  & 2020 & Modular pipeline                \\
A Survey of Autonomous Driving: Common Practices and Emerging Technologies \cite{survey_5}                                                  & Overall                  & 485  & 2020 &  Automated driving systems              \\
Self-driving cars: A survey \cite{survey_12}                                                                                           &   Overall                & 395  & 2021 &  Architecture      \\
Autonomous Cars: Research Results, Issues, and Future Challenges \cite{survey_3}                                                            &   Overall                & 291  & 2019 & Challenge               \\
Artificial Intelligence Applications in the Development of Autonomous Vehicles: A Survey \cite{survey_10}                                   &    Overall               & 107  & 2020 &  Emerging technologies               \\
Simultaneous localization and mapping: A survey of current trends in autonomous driving \cite{survey_loca_1}                                    &    Localization               & \textbf{489}  & 2017 &   SLAM     \\
A survey of the state-of-the-art localization techniques and their potentials for autonomous vehicle applications \cite{survey_loca_2}                                 &    Localization               & 433  & 2018 & Internet of Vehicles               \\
A review of visual-LiDAR fusion based simultaneous localization and mapping \cite{survey_loca_10}                                &    Localization               & 120  & 2020 & Autonomous navigation               \\
Vehicle dynamic state estimation: state of the art schemes and perspectives \cite{survey_loca_4}                                &    Localization               & 100  & 2018 & Pose estimation            \\
A review of recent advances in lane detection and departure warning system \cite{survey_static_2}                                                 & Static Detection  & \textbf{185}  & 2018 & Lane departure warning    \\
Advances in Vision-Based Lane Detection: Algorithms, Integration, Assessment, and Perspectives on ACP-Based Parallel Vision \cite{survey_static_3} &  Static Detection                 & 104  & 2018 & ACP parallel theory \\
Computer Vision for Autonomous Vehicles: Problems, Datasets and State of the Art \cite{survey_dynamic_13}                                            & Dynamic Detection & \textbf{532}  & 2020 &   Datasets              \\
Autonomous vehicle perception: The technology of today and tomorrow \cite{survey_dynamic_5}                                       & Dynamic Detection                  & 381  & 2018 &   Autonomous vehicle application  \\
Deep Multi-Modal Object Detection and Semantic Segmentation for Autonomous Driving: Datasets, Methods, and Challenges \cite{survey_dynamic_20}       &  Dynamic Detection                 & 331  & 2021 &  Fusion              \\
A Survey on 3D Object Detection Methods for Autonomous Driving Applications \cite{survey_dynamic_7}                                                &  Dynamic Detection                 & 230  & 2019 & 3D object detection         \\
Benchmarking Robustness in Object Detection: Autonomous Driving when Winter is Coming \cite{survey_dynamic_8}                                       &   Dynamic Detection                & 150  & 2019 & Benchmark datasets      \\
Object recognition and detection with deep learning for autonomous driving applications \cite{survey_dynamic_1}                                     &   Dynamic Detection                & 112  & 2017 & CNN and SVM      
       \\
Deep Learning-based Vehicle Behaviour Prediction For Autonomous Driving Applications: A Review \cite{survey_prediction_2}                                   &   Prediction                & \textbf{141}  & 2020 & Deep Learning      
       \\
A Literature Review on the Prediction of Pedestrian Behavior in Urban Scenarios \cite{survey_prediction_1}                                  &   Prediction                & 106  & 2018 &  Pedestrian Behavior     
       \\
A survey of motion planning and control techniques for self-driving urban vehicles \cite{survey_planning_1}                                   &  Planning                 & \textbf{1637}  & 2016 & Information sharing and coordination
       \\
Planning and Decision-Making for Autonomous Vehicles \cite{survey_planning_2}                                   &  Planning                 & 561  & 2018 & Fleet management
       \\
A Review of Motion Planning for Highway Autonomous Driving \cite{survey_planning_4}                                  &  Planning                 & 170  & 2020 &  Highway
       \\
Trajectory planning and tracking for autonomous overtaking: State-of-the-art and future prospects \cite{survey_planning_3}                                 &  Planning                 & 143  & 2018 &  High-speed driving 
       \\
Deep Reinforcement Learning for Autonomous Driving: A Survey \cite{survey_e2e_1}                                &  E2E                 & \textbf{143}  & 2018 & Reinforcement Learning
       \\
Driving Style Recognition for Intelligent Vehicle Control and Advanced Driver Assistance: A Survey \cite{survey_control_1}                               &  Control                 & \textbf{372}  & 2017 & Driving style
       \\
A Survey of Deep Learning Applications to Autonomous Vehicle Control \cite{survey_control_6}                              &  Control                 & 211  & 2021 & Vehicle control
       \\
Automated guided vehicle systems, state-of-the-art control algorithms and techniques \cite{survey_control_2}                             &  Control                 & 109  & 2019 & Automated guided vehicle system
       \\
The architectural implications of autonomous driving: Constraints and acceleration \cite{survey_system_1}                       &  System                 & \textbf{290}  & 2018 &  Hardware
       \\
Edge computing for autonomous driving: Opportunities and challenges \cite{survey_system_2}              &  System                 & 235  & 2019 & Connected  
       \\
Sensor technology in autonomous vehicles: A review \cite{survey_hardware_1}                          &  System              & 108  & 2018 & Sensor Fusion
       \\
Control of connected and automated vehicles: State of the art and future challenges \cite{survey_communication_4}                       &  Communication                 & \textbf{322}  & 2018 & Energy efficiency
       \\
A Survey of the Connected Vehicle Landscape—Architectures, Enabling Technologies, Applications,and Development Areas \cite{survey_communication_3}                        &  Communication                 & 283  & 2017 & Vehicle-to-everything
       \\
A Survey of Vehicle to Everything (V2X) Testing \cite{survey_communication_10}                        &  Communication                 & 167  & 2019 &  V2X application
       \\
A Survey of Intrusion Detection for In-Vehicle Networks \cite{survey_communication_11}                    &  Communication                 & 133  & 2019 & Controller area network
       \\
Autonomous vehicles that interact with pedestrians: A survey of theory and practice \cite{survey_human_machine_1}                    &  Interaction                 & \textbf{378}  & 2019 &  Pedestrian behavior
       \\
Collaborative vehicle routing: a survey \cite{survey_special_2}               &  Interaction                 & 213  & 2017 & Transportation
\\
\hline
\end{tabular}
\label{tabel:overall_key_surveys}
\end{sidewaystable*}

\section{Datasets}

The publicity of various kinds of autonomous driving datasets has made a substantial contribution to the advancement in this area, especially for perception and E2E planning tasks. KITTI \cite{dataset_KITTI} provides multiple computer vision tasks on urban roads in Germany. Cityscapes \cite{dataset_seg_CityScape}, BDD100K \cite{BDD100K}, Mapillary Vistas \cite{dataset_seg_Mapillary_Vistas} have released a number of data with segmentation masks. A*3D \cite{A*3D} enriches the collection scenes, such as the dark night, rainy and snowy.

Some automobile manufacturers publish datasets collected by their vehicles, including H3D \cite{H3D}, A2D2 \cite{A2D2} and the Ford Dataset \cite{Ford}. For more details, please check Table \ref{table:datasets}, which includes the number of frames, installed sensors, and covering tasks for each dataset. Readers could find the corresponding data by their mission. For E2E planning, the environment is more crucial for developers, and the simulation platform such as Carla \cite{CARLA}, Vissim \cite{fellendorf2010microscopic}, PerScan, AirSim \cite{shah2018airsim}, Udacity, Apollo, etc. could assist researchers to conduct experiments on planning and control.

\begin{table*}[]
\caption{The datasets on the autonomous driving$ ^1$}
\begin{threeparttable}
\begin{tabular}{lcp{0.15cm}p{0.15cm}p{0.15cm}p{0.15cm}p{0.15cm}p{0.15cm}p{0.15cm}p{0.15cm}p{0.15cm}p{0.15cm}p{0.15cm}p{0.15cm}p{0.15cm}p{0.15cm}p{0.15cm}p{0.15cm}p{0.15cm}p{0.15cm}p{0.15cm}p{0.15cm}p{0.15cm}p{0.15cm}p{0.15cm}p{0.15cm}p{0.20cm}}
\hline

\multicolumn{1}{c|}{\multirow{2}{*}{Dataset}} & \multicolumn{1}{c|}{\multirow{2}{*}{Frame}} & \multicolumn{7}{c|}{Sensors}                                  & \multicolumn{17}{c}{Task}                                                                                   \\ \cline{3-27} 
\multicolumn{1}{c|}{}                         & \multicolumn{1}{c|}{}                       & Li & Vi & Ra & GP & IM & Ca & \multicolumn{1}{c|}{Te} & Sc & Od & La & Dr & 2D & 3D & Di & OF & SF & PS & Se & Pa & De & Tr & Pr & Pl & E2E & HD\\ \hline

KITTI \cite{dataset_KITTI}                                         & 15K                                         & 1     & 2      & -     & 1   & 1   & - & -                        &     & \checkmark    &     &     & \checkmark    & \checkmark    & \checkmark   & \checkmark    & \checkmark    &        & \checkmark   & \checkmark   & \checkmark   & \checkmark   &     &     &     \\
CityScapes \cite{dataset_seg_CityScape}                                     & 25K                                         & -     & 2      & -     & 1   & 1   & - & 1                        &     &      &     &     & \checkmark    & \checkmark    &     &      &      &        & \checkmark   & \checkmark   &     &     &     &     &     \\
nuScenes \cite{nuScenes}                                     & 40K                                         & 1     & 6      & 5     & 1   & 1   & - & -                         &     &      &     &     & \checkmark    & \checkmark    &     &      &      & \checkmark      &     & \checkmark   &     & \checkmark   & \checkmark   &     &     \\
A2D2 \cite{A2D2}                                          & 12K                                         & 5     & 6      & -     & 1   & -   & - & -                       &     &      &     &     & \checkmark    & \checkmark    &     &      &      & \checkmark      & \checkmark   &     &     &     &     &     &     \\
Lyft L5 \cite{Lyft}                                      & -                                           & 3     & 7      & -     & -   & -   & -  & -                       &     &      &     &     &      & \checkmark    &     &      &      &        &     &     &     &     & \checkmark   &     &     \\
A*3D \cite{A*3D}                                        & 39K                                         & 1     & 2      & -     & -   & -   & -  & -                       &     &      &     &     & \checkmark    & \checkmark    &     &      &      &        &     &     &     &     &     &     &     \\
ApolloScape \cite{ApolloScape}                                  & 144K                                        &       &     &   &       &     &     &                          &     & \checkmark    & \checkmark   &     & \checkmark    & \checkmark    & \checkmark   &      &      & \checkmark      & \checkmark   &     & \checkmark   & \checkmark   &     & \checkmark   &     \\
BDD100K \cite{BDD100K}                                      & 100K                                        & -     & 1      & -     & 1   & 1   & - & -                        & \checkmark   &      &     & \checkmark   & \checkmark    &      &     &      &      &        & \checkmark   & \checkmark   &     & \checkmark   &     &     &     \\
H3D \cite{H3D}                                          & 27K                                         &1       & -   & -   &  1     & 1    & 1    &  -                        &     &      &     &     &      & \checkmark     &     &      &      &        &     &     &     & \checkmark    &     &     &     \\
Argoverse \cite{Argoverse}                                    & 22K                                         &2       &9    &-    &1       &-     &-     &-                          &     &      &     &     & \checkmark     &\checkmark      &\checkmark     &      &      &        &     &     & \checkmark    & \checkmark    & \checkmark    &     &   & \checkmark  \\
Mapillary Vistas \cite{dataset_seg_Mapillary_Vistas}                             & 25K   &1       &1   &-     &1       &-     &-     &-                          &     & \checkmark     &     &     & \checkmark     & \checkmark     &     &      &      & \checkmark       & \checkmark    & \checkmark    &\checkmark     &     &     &     &\checkmark      \\
Waymo Open \cite{Waymo}                                   & 200K                                        &5       &5    &-    &-       &-     &-     & -                         &     &      &     &     &\checkmark      &\checkmark      &     &      &      &\checkmark        &\checkmark     &\checkmark     &     &\checkmark     &     &     &     \\
Comma2k19 \cite{Comma2k19}                                    & 200K                                        &-       & 1  &-     & 1      &1     & 1    &    -                      &     & \checkmark   & \checkmark    &     &     &      &     &      &      &        &     &\checkmark    &     &     &     &     &     \\
Ford Dataset \cite{Ford}                                 & 200K                                        &4       &7    &-    &  1     &1     & -    &  -                        &     &\checkmark      &     &     & \checkmark     &  \checkmark    &     &      &      &        &     &     &     &     &     &     &     \\
PandaSet \cite{PandaSet}                                     & 16K                                         &2       &6    &-    &  1     &1     &     &  1                         &     &      & \checkmark     &  \checkmark    &      &      &     &      &  \checkmark     &        &     &     &     &     &     &     &     \\
ONCE \cite{ONCE}                                         & 1M                                          &1       & 7    &-   &-       &-     &-     &   -                       &     &      &     &     &  \checkmark    &\checkmark      &     &      &      &        &     &     &     &     &     &     &     \\
AutoMine \cite{AutoMine}                                     & 18K                                         &1       &2    & -   & 1      &1     &-     & -                         &     & \checkmark     &     &     & \checkmark     & \checkmark     &     &      &      &        &     &     &     &     &     &     &    
\\ \hline
\end{tabular}
      \begin{tablenotes}
        \footnotesize
        \item[1] Li-LiDAR, Vi-Vision, Ra-Radar, GP-Global Positioning System, IM-Inertial Measurement Unit, Ca-CAN data, Te-Temperature data, Sc-Scene Classification, Od-Odometry, La-Lane Detection, Dr-Driveable Detection, 2D-2D Object Detection, 3D-3D Object Deteciton, Di-Disparity, OF-Optical Flow Estimation, SF-Scene Flow Estimation, PS-Point Segmentation, Se-Semantic Segmentation, Pa-Panoptic Segmentation, De-Depth Estimation, Tr-Tracking, Pr-Prediction, Pl-Planning, E2E-End-to-End, HD-High Definition map.
      \end{tablenotes}
\label{table:datasets}
\end{threeparttable}
\end{table*}
\section{Perspectives and  Future}

\subsection{Independent Tasks:}

\subsubsection{\textbf{Perception}}

Perception is the upstream aspect of autonomous driving systems, and the results of which will heavily influence downstream tasks including planning and motion control. Combined with limited computational resources and time, perception models need to be accurate, robust, and fast. A number of teams have achieved competitive results in academic research on perception, but researchers still need to continue to improve the performance of their models until they could cover the full scene, which is the fundamental characteristic of mass production. We summarise a few possible future research directions as follows: 1) The early fusion strategies \& universal structures for multiple sensors. 2) Lifting the 2D to 3D  detection adopting effective transfer structures. 3) Making IVs have the capability of automated inference. 4) Developing self-supervised strategies and reducing the relay on huge data. 5) Exploring the cooperative perception and making a dense connection to the following tasks.

\subsubsection{\textbf{Planning}}
Trajectory planning technique alone is not the bottleneck of an IV. Despite this, the planning module deserves to consider the limitations in the upstream/downstream modules so that the entire driving performance is improved. The following few aspects are an outlook on some possible future directions: 1) Safe planning for imperfect perception data. 2) Balance of solution quality and speed. 3) Performance consistency in switching between different planners. 4) Interpretability enhancement for a learning-based planner \cite{teng2023motion}.

\subsubsection{\textbf{Control}}

The motion control technology of IVs has made remarkable progress.  However, due to the complex longitudinal and lateral dynamics of the vehicle, mutual coupling performance objectives, and the wide application of advanced communication technology, there are still many important and unsolved problems in the research of IV motion control that need to be explored and recognized. The following is a preliminary outlook on its possible development directions: 1) Coordinated control method of longitudinal and lateral motion of IV under random uncertainty and delay conditions. 2) Multi-performance objective global optimization technology for IV motion control. 3) Theory and method of IV cooperative control in the Internet of Vehicles environment. 4) Fault tolerant method for control systems. 5) Piratical application of control systems in a real traffic environment.

\subsubsection{\textbf{Testing}}

Testing is a crucial process before the mass production of IVs. Test vehicles require to complete a series of driving tasks with various difficulties in testing areas or private roads. The purpose of this process is to locate the remaining problems of IVs, to provide the last opportunity to modify the program and to reduce the accident rate of IVs on public roads. For future research on IVs testing, researchers could 1) introduce a novel evaluation criterion on thinking rationally; 2) develop the evaluation criteria for virtual simulation testing; 3) attempt to narrow the gap between the real and virtual testing scenarios \cite{test_final_tiv_1, test_final_tiv_2}.

\subsubsection{\textbf{Human Behaviors}}
The increase of autonomy for commercial passenger vehicles will not reduce the necessity of human behaviors and human factors issues but may increase the complexity of these problems. The responsibility of ensuring a safe, comfortable, and pleasant journey is extremely heavy for IVs. Future work for HMI systems on IVs should further focus on the development of mutual understanding and mutual trust mechanisms, ensuring communication transparency and efficiency with both onboard and surrounding users. The personalized and human-centered design approach should be highlighted to guarantee the IVs are also able to understand user characteristics and personalities in case to interact with humans more effectively. Meanwhile, security, privacy, and ethics issues are also expected to be carefully considered \cite{human-machine_1, human-machine_3, human-machine_5, human-machine_6, human-machine_15, human-machine_21}.

\subsection{Ethics on Autonomous Driving:}


\subsubsection{\textbf{Normative Ethics}}


The normative ethical issues centre around the moral dilemmas where an IV has to make a choice between alternatives that will inevitably result in the sacrifice of human lives \cite{e2,e3,e5}. 
One example of adapting the trolley problem of IV is given by Bonnefon et al. \cite{e2}, who designed several delicated accident scenarios where an IV has to make decisions between scarifying pedestrians or passengers and surveyed the choices held by participants in the US. Results of the survey suggested that most participants wanted other people to buy IVs prioritising saving the most lives in the accidents. 
The survey mechanism was expanded and developed to an online experimental platform known as the ``Moral Machine Experiment" \cite{e3} to explore the moral dilemmas faced by IVs from a global perspective. The data helped identify three strong preferences: the preference for sparing human lives, that for sparing more lives, and that for sparing young lives.

In a separate study, Morita and Managi \cite{e5} surveyed the existence of the social dilemma in Japan and found that the result is broadly similar to those obtained in the USA \cite{e3}. Of particular note is that participants in the US expressed generally stronger preferences for self-protective IVs when travelling with family over riding alone, while those in Japan did not demonstrate such inclination. This difference, argued by the authors, was due to the cultural difference between the two countries. While discussions referring to the trolly problem in the context of IV moral decision-making have been comprehensive and rich, several researchers have expressed concerns that the IVs moral dilemmas were overstressed. Cunneen et al. \cite{e8} argued that framing the ethical impact of IVs in terms of the trolley-problem-like dilemmas was misleading, while more realistic ethical framings should focus on the present and near-future technologies including HMI, machine perception, and data privacy, etc. This attitude was shared by Lundgren \cite{e9}, who also questioned the methodologies in which the discussions on the IVs’ trolley problem were extended.

\subsubsection{\textbf{Environmental and Public Health Ethics}}


A consensus has been achieved that the introduction of IVs will raise issues associated with environmental and public health ethics. IVs could benefit the environment by, e.g. optimising energy efficiency and emissions of individual vehicles and reducing traffic congestions caused by collisions \cite{e10}. Meanwhile, IVs could bring harms to the environment \cite{e11,e12,e13,e14}. The convenience and accessibility of IVs could unlock the additional travel demand from people who bear unnecessary travel needs \cite{e11}, and the increased travel demand would in turn increased the Vehicle-Miles-Travelled (VMT) \cite{e12}, and result in higher levels of noise and ElectroMagnetic Fields (EMF), both of which contribute to adverse health effects \cite{e13}.

\subsubsection{\textbf{Business Ethics}}


Another two heated debate topics raised by IVs are liability and privacy, both of which are primarily targeted on the IV industry and thus fall into the theme of business ethics. Unlike most conventional vehicle accidents, accidents associated with an IV could cause controversial legal issues regarding the apportion of liability among the industrial stakeholders of the IV technology \cite{e16}. What could make the issues even worse is that these stakeholders might not even be able to predict the behaviour of an IV due to the inherent unpredictability of the machine-learning-based algorithms \cite{e18}. In conjunction, serious privacy risks could arise as the IV industry prosper – it would become increasingly uncomplicated for the industrial stakeholders to access the IV users’ personal information \cite{e16,e18}.

\subsection{Future Directions:}



\subsubsection{\textbf{Human-Machine Hybrid Intelligence}}

The relationship between human and IVs is not independent. On the contrary, both are coexistent and mutually reinforcing. Human intelligence is the mentor of machine intelligence, and the latter will learn problem-solving strategies from human behaviors, so as to improve the reliability of intelligent systems \cite{Hybrid_1, hybrid_2}. 

At the American Association for Artificial Intelligence (AAAI) conference in 2018, the conference president gave a presentation named ``Challenge of human aware Artificial Intelligence (AI) systems", which pointed out the challenges we face in the development of AI systems. The presentation suggests that the purpose of AI is to augment the human labor, so in order to collaborate with AI systems, it is necessary to design them with human awareness and to build models of Human in the Loop (HITL). AD is one of typical AI systems, and it needs to combine AI algorithms with human involvement, and a HITL approach will enhance the ability of IVs to handle complex difficulties.

\subsubsection{\textbf{Parallel Intelligence in Autonomous Driving}}

Human drivers are mostly capable to detect important information from the surrounding environment and thus make rational decisions. However, this type of capability relies on a large amount of knowledge. A parallel simulation platform based parallel intelligence can greatly enrich the perception data through data enhancement in virtual scenarios. It creates abundant corner cases and diverse weather conditions to enhance detection \& planning capabilities in virtual scenarios \cite{parallel_1, parallel_2, parallel_3, parallel_5, parallel_final_1, parallel_final_tiv_1}. In addition, through correlation guidance between virtual and realistic scenarios, models trained in virtual environments can be deployed into real IVs to improve the capability of models on urban roads.



\subsubsection{\textbf{From Scenario Engineering to Scenario Intelligence}}

Nowadays, the scenario datasets store information with different formats and standards, and without effective indexing. Thus these datasets are sparsely annotated and difficult to reuse. The purpose of scenario intelligence is to uniform the description methods \& rules \cite{chenlong_tip, anno_1, scene_final_1}. Through scenario intelligence, IVs will be able to adapt to various road conditions and driving environments, improving the intelligence level, which is one of the crucial technologies to achieve L5 AD level in the future.
\section{conclusion}
In view of the large number of surveys lacking systematic summaries and macroscopic perspectives, we have made a comprehensive summary of AD and IVs. This article is Part 1 of our work. In this paper, we review the development of AD and introduce an SoS on milestone research in AD and IVs. We collect $122$ surveys into $18$ categories by research areas and analyse them. Datasets on AD are summarized to assist researchers to select the suitable data as fast as possible. In addition, we have pointed out the research perspectives, ethics and a few future directions on AD. This SoS offers horizontal as well as vertical research on various topics in AD.

\small\bibliographystyle{IEEEtran}
\bibliography{body/mylib.bib}


\section{Biography Section}

\begin{IEEEbiography}[{\includegraphics[width=1in,height=1.25in,clip,keepaspectratio]{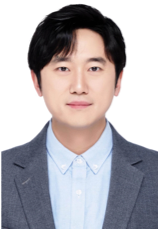}}]{Long Chen} (Senior Member, IEEE) received the Ph.D. degree from Wuhan University in 2013, he is currently a Professor with State Key Laboratory of Management and Control for Complex Systems, Institute of Automation, Chinese Academy of Sciences, Beijing, China. His research interests include autonomous driving, robotics, and artificial intelligence, where he has contributed more than 100 publications. He serves as an Associate Editor for the IEEE Transaction on Intelligent Transportation Systems, the IEEE/CAA Journal of Automatica Sinica, the IEEE Transaction on Intelligent Vehicle and the IEEE Technical Committee on Cyber-Physical Systems.
\end{IEEEbiography}

\begin{IEEEbiography}[{\includegraphics[width=1in,height=1.25in,clip,keepaspectratio]{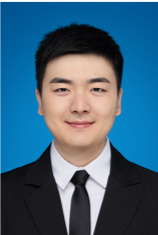}}]{Yuchen Li} received the B.E. degree from the University of Science and Technology Beijing in 2016, and the M.E. degrees from Beihang University in 2020. He is pursuing the Ph.D. degree in Hong Kong Baptist University. He is a intern in Waytous. His research interest covers computer vision, 3D object detection, and autonomous driving.
\end{IEEEbiography}

\begin{IEEEbiography}[{\includegraphics[width=1in,height=1.25in,clip,keepaspectratio]{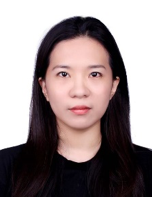}}]{Chao Huang} Chao Huang received her B.S. degree in Control Engineering from the China University of Petroleum, Beijing, and a Ph.D. degree from the University of Wollongong, Australia in 2012 and 2018, respectively. She is currently a research assistant professor at the Department of Industrial and Systems Engineering, The Hong Kong Polytechnic University. Her research interests are human-machine collaboration, fault-tolerant control, mobile robot (EV, UAV), and path planning and control.  She has served on program committees and helped to organize special issues on Sensors, Machines, and Aerospace. She has served as review editor of Frontiers in Mechanical Engineering and associate editor of IEEE Transactions on Intelligent Vehicles.
\end{IEEEbiography}

\begin{IEEEbiography}[{\includegraphics[width=1in,height=1.25in,clip,keepaspectratio]{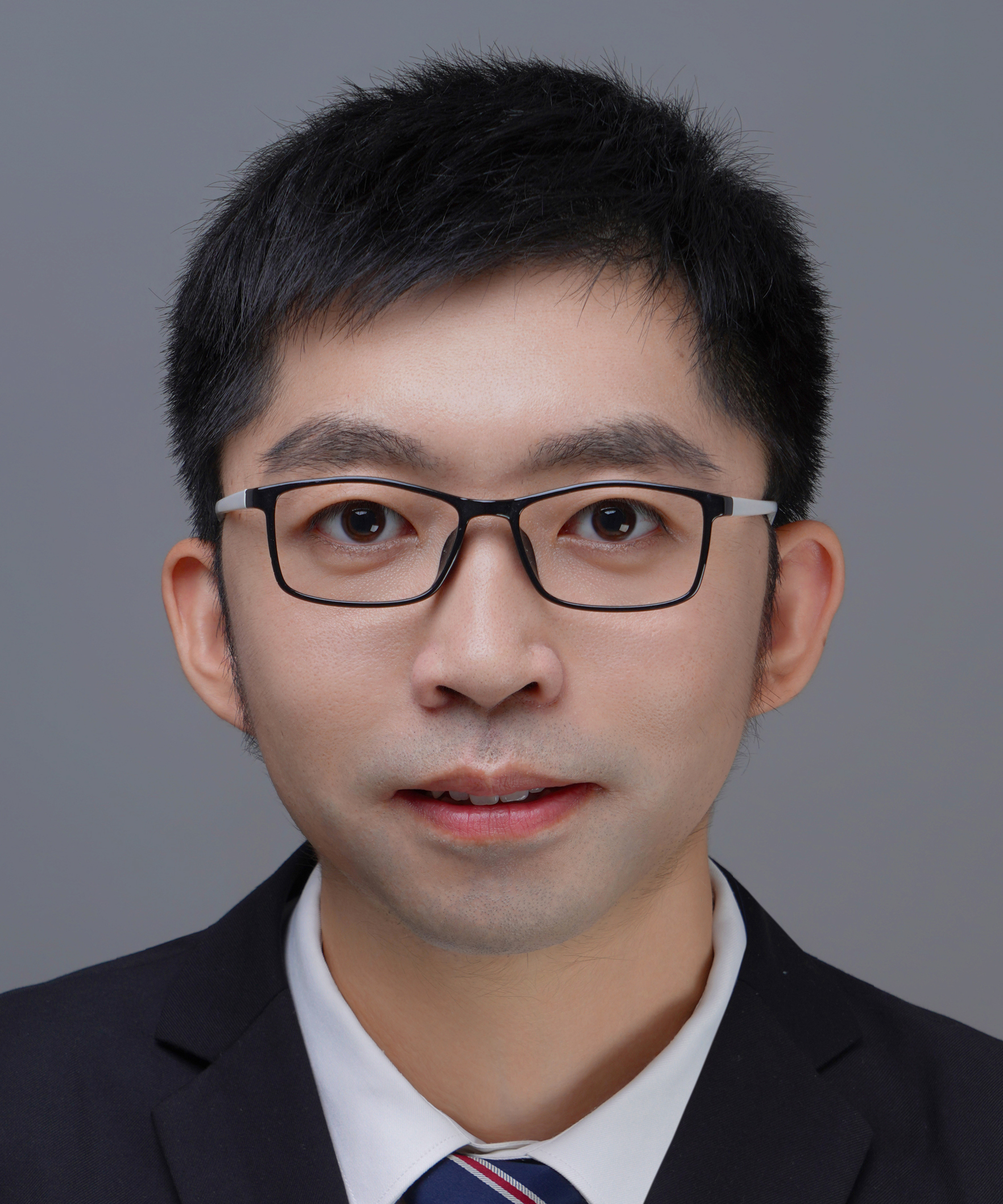}}]{Bai Li} (SM’13–M’18) received his B.S. degree in 2013 from Beihang University, China, and the Ph.D. degree in 2018 from the College of Control Science and Engineering, Zhejiang University, China. He is currently an associate professor in the College of Mechanical and Vehicle Engineering, Hunan University, Changsha, China. Before joining Hunan University, he was a research engineer of JD.com Inc., Beijing, China from 2018 to 2020. Prof. Li was the first author of more than 60 journal/conference papers and two books in numerical optimization, optimal control, and trajectory planning. He was a recipient of the International Federation of Automatic Control (IFAC) 2014–2016 Best Journal Paper Prize. He is an Associate Editor of IEEE TRANSACTIONS ON INTELLIGENT VEHICLES.
\end{IEEEbiography}

\begin{IEEEbiography}[{\includegraphics[width=1in,height=1.25in,clip,keepaspectratio]{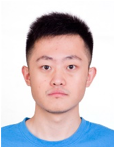}}]{Yang Xing} received his Ph. D. degree from Cranfield University, UK, in 2018. He joined Cranfield University as a Lecturer in Applied AI for Engineering in 2021. Before that, he finished his post-doctoral fellowship at the University of Oxford and Nanyang Technological University, respectively.  His research interests include machine learning, driver behaviours, intelligent multi-agent collaboration, and intelligent/autonomous vehicles. He received the IV2018 Best Workshop/Special Issue Paper Award. Dr Xing serves as an associate editor for IEEE Transactions on Intelligent Vehicle. He was also a Guest Editor for IEEE Internet of Things Journal, and IEEE ITS Magazine, etc. 
\end{IEEEbiography}

\begin{IEEEbiography}[{\includegraphics[width=1in,height=1.25in,clip,keepaspectratio]{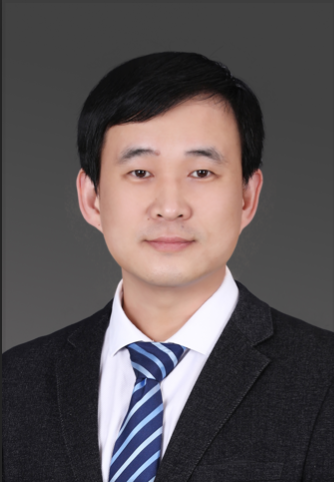}}]{Daxin Tian}
 (Senior Member, IEEE) received the B.S., M.S., and Ph.D. degrees in computer science from Jilin University, Changchun, China, in 2002, 2005, and 2007, respectively. He is currently a professor in the School of Transportation Science and Engineering, Beihang University, Beijing, China. He is an IEEE Intelligent Transportation Systems Society Member and an IEEE Vehicular Technology Society Member. His current research interests include mobile computing, intelligent transportation systems, vehicular ad hoc networks, and swarm intelligent.
\end{IEEEbiography}

\begin{IEEEbiography}[{\includegraphics[width=1in,height=1.25in,clip,keepaspectratio]{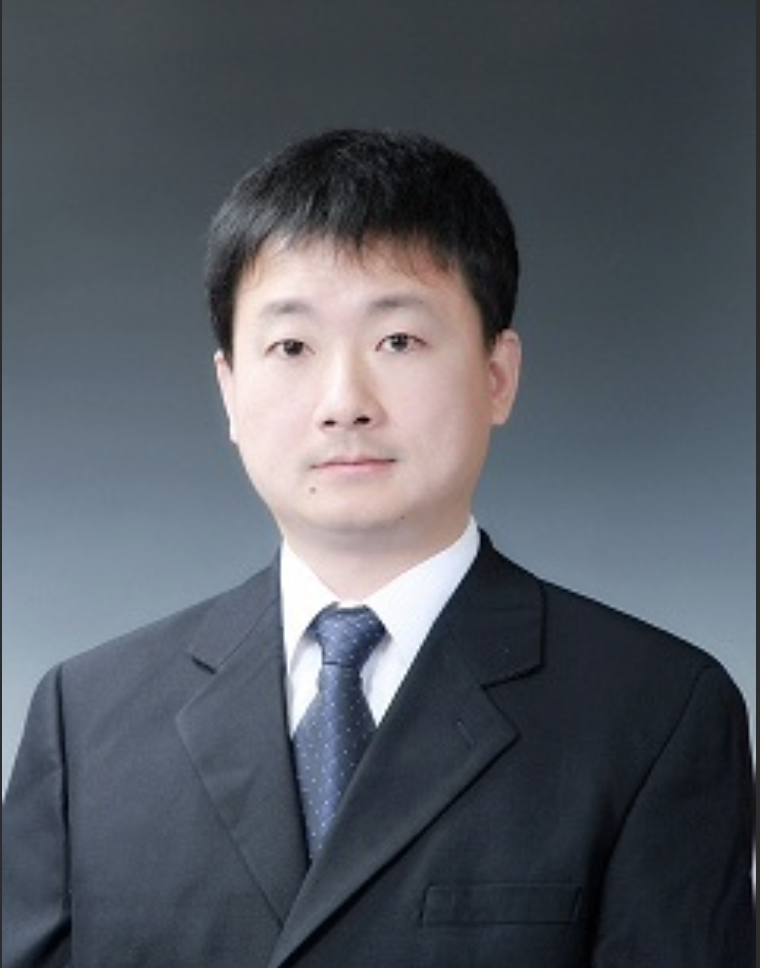}}]{Li Li} (Fellow, IEEE) is a Professor at Tsinghua University. He has been engaged in scientific research in the fields of intelligent transportation and intelligent vehicles. He has published more than 120 SCI search papers as the first or corresponding author. He was a member of the Editorial Advisory Board for Transportation Research Part C: Emerging Technologies, a member of the Editorial Board of Transport Reviews and ACTA Automatica Sinica. He serves as Associate Editors for the IEEE Transactions on Intelligent Transportation Systems and IEEE Transactions on Intelligent Vehicles. He is a Fellow of IEEE and a Fellow of CAA.
\end{IEEEbiography}

\begin{IEEEbiography}[{\includegraphics[width=1in,height=1.25in,clip,keepaspectratio]{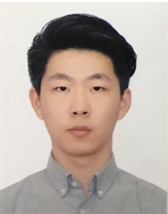}}]{Zhongxu Hu} received a mechatronic Ph.D. degree from the Huazhong University of Science and Technology of China, in 2018. He was a senior engineer at Huawei. He is currently a research fellow within the Department of Mechanical and Aerospace Engineering of Nanyang Technological University in Singapore. His current research interests include human-machine interaction, computer vision, and deep learning applied to driver behavior analysis and autonomous vehicles in multiple scenarios. Dr. Hu served as a Lead Guest Editor for Computational Intelligence and Neuroscience, an Academic Editor/Editorial Board for Automotive Innovation, Journal of Electrical and Electronic Engineering, and is also an active reviewer for IEEE Transactions on Intelligent Transportation Systems, IEEE Transactions on Industrial Electronics, IEEE Intelligent Transportation Systems Magazine, Journal of Intelligent Manufacturing, and Journal of Advanced Transportation et al.
\end{IEEEbiography}

\begin{IEEEbiography}[{\includegraphics[width=1in,height=1.25in,clip,keepaspectratio]{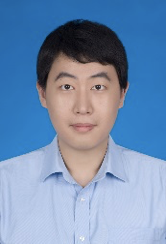}}]{Xiaoxiang Na}received the B.Sc. and M.Sc. degrees in automotive engineering from the College of Automotive Engineering, Jilin University, China, in 2007 and 2009, respectively. He received the Ph.D. degree in driver-vehicle dynamics from the Department of Engineering, University of Cambridge, U.K. in 2014. He is currently a Senior Research Associate with the Centre for Sustainable Road Freight, University of Cambridge, and a Borysiewicz Interdisciplinary Fellow at the University of Cambridge. His main research interests include driver-vehicle dynamics, operational monitoring of road freight vehicles, and vehicle energy performance assessment.
\end{IEEEbiography}

\begin{IEEEbiography}[{\includegraphics[width=1in,height=1.25in,clip,keepaspectratio]{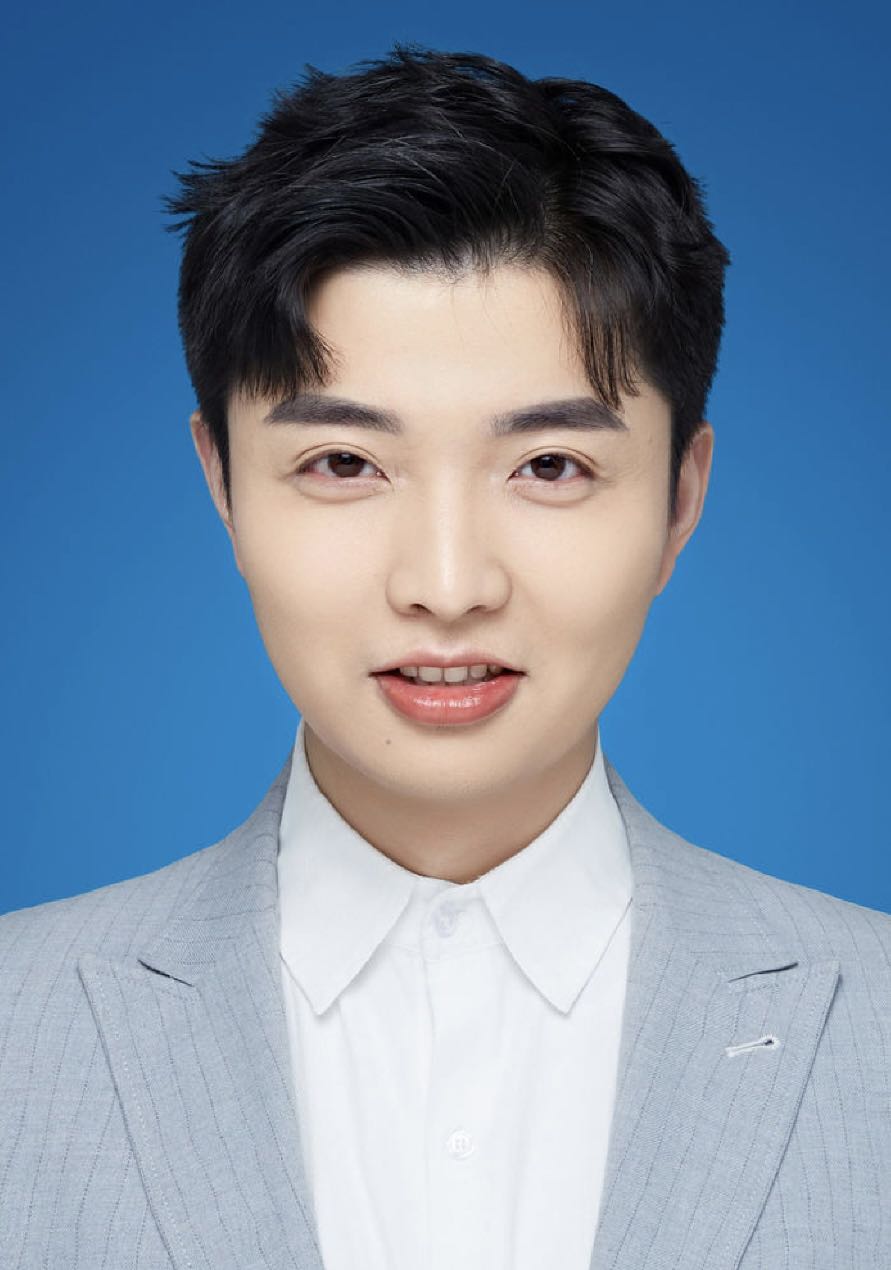}}]{Zixuan Li} received the B.E. degree from Anhui University in 2017,and the M.E. degree from the University of Chinese Academy of Sciences in 2021. He is a intern in Waytos. His research interest cover computer vision ,communication engineering and autonomous driving.
\end{IEEEbiography}

\begin{IEEEbiography}[{\includegraphics[width=1in,height=1.25in,clip,keepaspectratio]{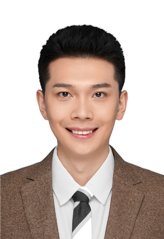}}]{Siyu Teng} received M.S. degree from Jilin University in 2021. Now he is a PhD Student at Department of Computer Science, Faculty of Science, Hong Kong Baptist University. His main interests are parallel planning, end-to-end autonomous driving and interpretable deep learning.
\end{IEEEbiography}

\begin{IEEEbiography}[{\includegraphics[width=1in,height=1.25in,clip,keepaspectratio]{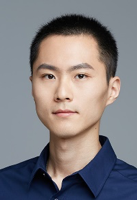}}]{Chen Lv} (Senior Member, IEEE) is a Nanyang Assistant Professor at School of Mechanical and Aerospace Engineering, Nanyang Technological University, Singapore. He received his PhD degree at Department of Automotive Engineering, Tsinghua University, China, with a joint PhD at UC Berkeley. His research focuses on intelligent vehicles, automated driving, and human-machine systems, where he has contributed 2 books, over 100 papers, and obtained 12 granted patents. He serves as Associate Editor for IEEE T-ITS, IEEE TVT, and IEEE T-IV. He received many awards and honors, selectively including the Highly Commended Paper Award of IMechE UK in 2012, Japan NSK Outstanding Mechanical Engineering Paper Award in 2014, Tsinghua University Outstanding Doctoral Thesis Award in 2016, IEEE IV Best Workshop/Special Session Paper Award in 2018, Automotive Innovation Best Paper Award in 2020, the winner of the Waymo Open Dataset Challenges at CVPR 2021, and Machines Young Investigator Award in 2022.
\end{IEEEbiography}

\begin{IEEEbiography}[{\includegraphics[width=1in,height=1.25in,clip,keepaspectratio]{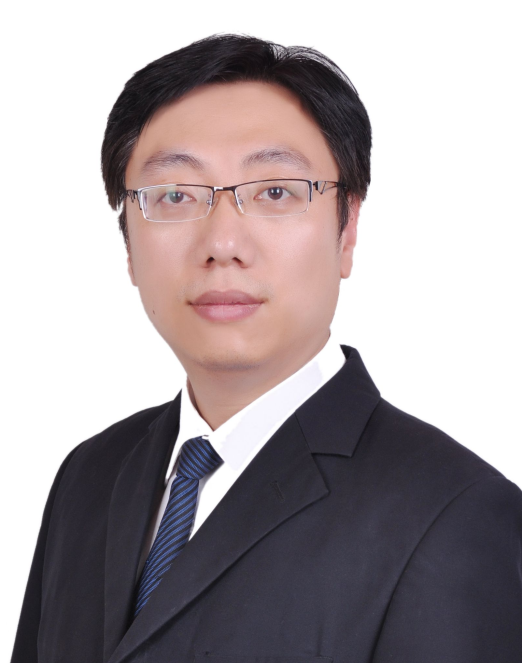}}]{Jinjun Wang} received the B.E. and M.E. degrees from Huazhong University of Science and Technology, China, in 2000 and 2003, respectively, and the Ph.D. degree from Nanyang Technological University, Singapore, in 2006. From 2006 to 2009, he was with NEC Laboratories America, Inc., as a Research Scientist, and from 2010 to 2013, he was with Epson Research and Development, Inc., as a Senior Research Scientist. He is currently a Professor with Xi’an Jiaotong University. His research interests include pattern classification, image/video enhancement and editing, content-based image/video annotation and retrieval, and semantic event detection.
\end{IEEEbiography}

\begin{IEEEbiography}[{\includegraphics[width=1in,height=1.25in,clip,keepaspectratio]{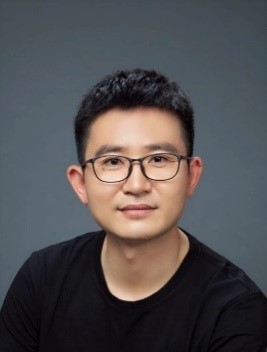}}]{Dongpu Cao} (Senior Member, IEEE) received the Ph.D. degree from Concordia University, Canada, in 2008. He is the Canada Research Chair in Driver Cognition and Automated Driving, and currently an Associate Professor and Director of Waterloo Cognitive Autonomous Driving (CogDrive) Lab at University of Waterloo, Canada. His current research focuses on driver cognition, automated driving and cognitive autonomous driving. He has contributed more than 200 papers and 3 books. He received the SAE Arch T. Colwell Merit Award in 2012, and three Best Paper Awards from the ASME and IEEE conferences. Dr. Cao serves as an Associate Editor for IEEE TRANSACTIONS ON VEHICULAR TECHNOLOGY, IEEE TRANSACTIONS ON INTELLIGENT TRANSPORTATION SYSTEMS, IEEE/ASME TRANSACTIONS ON MECHATRONICS, IEEE TRANSACTIONS ON INDUSTRIAL ELECTRONICS, IEEE/CAA JOURNAL OF AUTOMATICA SINICA and ASME JOURNAL OF DYNAMIC SYSTEMS, MEASUREMENT AND CONTROL. He was a Guest Editor for VEHICLE SYSTEM DYNAMICS and IEEE TRANSACTIONS ON SMC: SYSTEMS. He serves on the SAE Vehicle Dynamics Standards Committee and acts as the Co-Chair of IEEE ITSS Technical Committee on Cooperative Driving.
\end{IEEEbiography}

\begin{IEEEbiography}[{\includegraphics[width=1in,height=1.25in,clip,keepaspectratio]{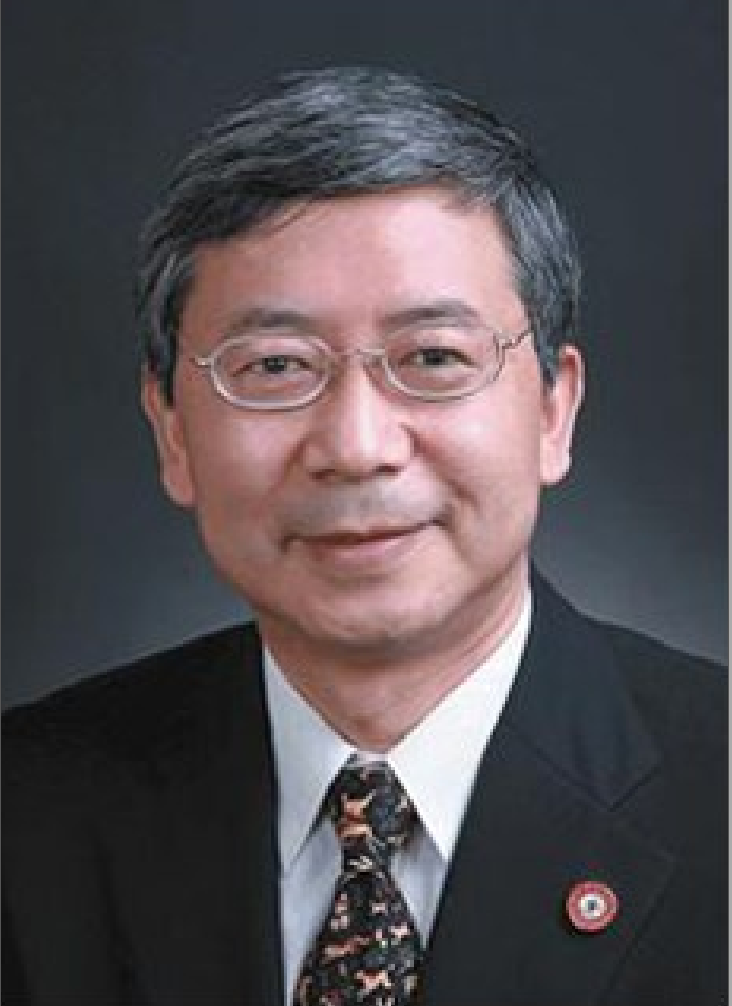}}]{Nanning Zheng} (SM93-F06) (Fellow, IEEE) graduated from the Department of Electrical Engineering, Xian Jiaotong University, Xian, China, in 1975, and received the M.S. degree in information and control engineering from Xian Jiaotong University in 1981 and the Ph.D. degree in electrical engineering from Keio University, Yokohama, Japan, in 1985. He jointed Xian Jiaotong University in 1975, and he is currently a Professor and the Director of the Institute of Artificial Intelligence and Robotics, Xian Jiaotong University. His research interests include computer vision, pattern recognition and image processing, and hardware implementation of intelligent systems. Dr. Zheng became a member of the Chinese Academy of Engineering in 1999, and he is the Chinese Representative on the Governing Board of the International Association for Pattern Recognition.
\end{IEEEbiography}

\begin{IEEEbiography}[{\includegraphics[width=1in,height=1.25in,clip,keepaspectratio]{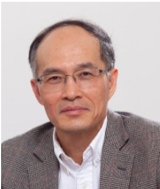}}]{Fei-Yue Wang} (Fellow, IEEE) received the Ph.D. degree in computer and systems engineering from the Rensselaer Polytechnic Institute, Troy, NY, USA, in 1990. He joined The University of Arizona, Tucson, AZ, USA, in 1990, and became a Professor and the Director of the Robotics and Automation Laboratory and the Program in Advanced Research for Complex Systems. In 1999, he founded the Intelligent Control and Systems Engineering Center, Institute of Automation, Chinese Academy of Sciences (CAS), Beijing, China, under the support of the Outstanding Chinese Talents Program from the State Planning Council, and in 2002, was appointed as the Director of the Key Laboratory of Complex Systems and Intelligence Science, CAS, and Vice President of Institute of Automation, CAS, in 2006. In 2011, he became the State Specially Appointed Expert and the Founding Director of the State Key Laboratory for Management and Control of Complex Systems. He has been the Chief Judge of Intelligent Vehicles Future Challenge since 2009 and Director of China Intelligent Vehicles Proving Center, Changshu since 2015. He is currently the Director of Intel’s International Collaborative Research Institute on Parallel Driving with CAS and Tsinghua University, Beijing, China. His research interests include methods and applications for parallel intelligence, social computing, and knowledge automation. He is a Fellow of INCOSE, IFAC, ASME, and AAAS. In 2007, he was the recipient of the National Prize in Natural Sciences of China, numerous best papers awards from IEEE Transactions, and became an Outstanding Scientist of ACM for his work in intelligent control and social computing. He was the recipient of the IEEE ITS Outstanding Application and Research awards in 2009, 2011, and 2015, respectively, IEEE SMC Norbert Wiener Award in 2014, and became the IFAC Pavel J. Nowacki Distinguished Lecturer in 2021. Since 1997, he has been the General or Program Chair of more than 30 IEEE, INFORMS, IFAC, ACM, and ASME conferences. He was the President of the IEEE ITS Society from 2005 to 2007, IEEE Council of RFID from 2019 to 2021, Chinese Association for Science and Technology, USA, in 2005, American Zhu Kezhen Education Foundation from 2007 to 2008, Vice President of the ACM China Council from 2010 to 2011, Vice President and the Secretary General of the Chinese Association of Automation from 2008 to 2018, Vice President of IEEE Systems, Man, and Cybernetics Society from 2019 to 2021. He was the Founding Editor-in-Chief (EiC) of the International Journal of Intelligent Control and Systems from 1995 to 2000, IEEE ITS Magazine from 2006 to 2007, IEEE/CAA JOURNAL OF AUTOMATICA SINICA from 2014 to 2017, China’s Journal of Command and Control from 2015 to 2021, and China’s Journal of Intelligent Science and Technology from 2019 to 2021. He was the EiC of the IEEE INTELLIGENT SYSTEMS from 2009 to 2012, IEEE TRANSACTIONS ON INTELLIGENT TRANSPORTATION SYSTEMS from 2009 to 2016, IEEE TRANSACTIONS ON COMPUTATIONAL SOCIAL SYSTEMS from 2017 to 2020. He is currently the President of CAA’s Supervision Council and new EiC of the IEEE TRANSACTIONS ON INTELLIGENT VEHICLES.
\end{IEEEbiography}

\vfill

\end{document}